# OPTIMIZING FACE RECOGNITION USING PCA

Manal Abdullah[1], Majda Wazzan[1] and Sahar Bo-saeed[2]

[1] Faculty of Computer Sciences and Information Technology, King Abdulaziz University, Jeddah, KSA

maaabdullah@kau.edu.sa    majda.wazzan@gmail.com

[2] College of Computer and Information Sciences, King Saud University, Riyadh, KSA

sahablight@yahoo.com

## ABSTRACT

*Principle Component Analysis PCA is a classical feature extraction and data representation technique widely used in pattern recognition. It is one of the most successful techniques in face recognition. But it has drawback of high computational especially for big size database. This paper conducts a study to optimize the time complexity of PCA (eigenfaces) that does not affects the recognition performance. The authors minimize the participated eigenvectors which consequently decreases the computational time. A comparison is done to compare the differences between the recognition time in the original algorithm and in the enhanced algorithm. The performance of the original and the enhanced proposed algorithm is tested on face94 face database. Experimental results show that the recognition time is reduced by 35% by applying our proposed enhanced algorithm. DET Curves are used to illustrate the experimental results.*

## 1. INTRODUCTION

Face recognition is one of the important challenges in appearance-based pattern recognition field. This technology has emerged as an attractive solution to address many new needs for identification and verification of identity. It is the more familiar functionality of visual surveillance systems. It has received significant attention for decades due to its numerous potential applications. Some of these applications are national security, low enforcement, surveillance, public safety field. There are many factors affect the face recognition performance like facial expression, illumination, pose, cluttered background or occlusion.

The process begins with face detection and extraction from the larger image, then normalizing the probe image so that it is in the same format (size, rotation, etc.) as the images in the database. The normalized face image is then passed to the recognition phase.

Face recognition can typically be used for verification or identification. In verification an individual is already enrolled in the reference database or gallery i.e. it is a one-to-one matching task whereas in identification, a probe image is matched with a biometric reference in the gallery i.e. it represents a one-to-many problem.

There are two outcomes: the person is not recognized or the person is recognized. Two recognition mistakes may occur: false reject (FR) which indicates a mistake that occur when the system reject a known person, false accept (FA) which indicates a mistake in accepting a claim when it is in fact false.

In the past many years, there are a plenty of work has been done in face recognition and have achieved success in real application. We can divide these algorithms into two main approaches: two dimensional (2D) approaches and three dimensional (3D) approaches. Mainly, the traditional 2D approaches are divided into six algorithms: eigenfaces (PCA), fisherfaces or linear discriminant analysis (LDA), independent component analysis (ICA), support vector machine (SVM), neural network and hidden markov model (HMM) [1].

The 3D face recognition approaches become more popular. This technique uses 3D sensors to capture information about the shape of a face. It can be divided in to two main categories: 3D face reconstruction and 3D pose estimation. A third kind of approaches is hybrid approach which combines 2D with 3D approaches together.

Despite the used algorithm, facial recognition can be decomposed into four phases: pre-processing phase, segmentation or localization, feature extraction phase and recognition phase.
One of the most popular algorithms is principal component analysis (PCA) [2]. In PCA, the probe and gallery images must be the same size. Therefore, a normalization is needed to lineup the eyes and the mouths across all images. Each image is treated as one vector. All images of the training set are stored in a single matrix T and each row in the matrix represents an image. The average image has to be calculated and then subtracted from each original image in T. Then calculate the eigenvectors and eigenvalues of the covariance matrix S. These eigenvectors are called eigenfaces. The eigenfaces is the result of the reduction in dimensions which removes the un-useful information and decomposes the face structure into the uncorrelated components (eigenfaces). Each image may be represented as a weighted sum of the eigenfaces. A probe image is then compared against the gallery by measuring the distance between their represent vectors.

In this paper the authors attempt to enhance the performance of the PCA by minimizing the eigenvectors which consequently decrease the computational time without greatly affect the recognition accuracy. Experiments are based on face94 [3] face database and show that the distances does not change after applying our proposed method and the false acceptance rate FAR can be decreased. The performance of the proposed algorithm is tested on face94 face database, and the obtained results show an improvement in performance of the proposed algorithm as compared to the same with PCA method.

The rest of this paper is organized as follows: in Section 2 we give a background illustrating the popular face recognition algorithm PCA (eigenfaces). Section 3, demonstrates the recent and related work in face recognition field. Section 4, we present our work. In Section 5 we display the conducted experiments, Section 6 demonstrates the experiment results and Section 7 concludes this work.

## 2. PRINCIPLE COMPONENT ANALYSIS

This section describes the eigenfaces approach [2]. This approach for face recognition aims to decompose face images into small set of characteristic feature images called eigenfaces which used to represent both existing and new faces.

The training database consists of M images which is same size. The images are normalized by converting each image matrix to equivalent image vector $\Gamma_i$. The training set matrix $\Gamma$ is the set of image vectors with

$$\text{Training set } \Gamma = [\Gamma_1 \, \Gamma_2 \, ….. \, \Gamma_M] \quad (1)$$

The mean face ($\psi$) is the arithmetic average vector as given by:

$$\psi = \frac{1}{M}\sum_{i=1}^{M} \Gamma_i \quad (2)$$

The deviation vector for each image $\Phi_i$ is given by:

$$\Phi_i = \Gamma_i - \psi \quad i = 1,2,…M \quad (3)$$

Consider a difference matrix A= [ $\Phi_1$, $\Phi_2$, ……. $\Phi_M$ ] which keeps only the distinguishing features for face images and removes the common features. Then eigenfaces are calculated by find the Covariance matrix C of the training image vectors by:

$$C = A.A^T \quad (4)$$

Due to large dimension of matrix C, we consider matrix L of size ($M_t$ X $M_t$) which gives the same effect with reduces dimension.

The eigenvectors of C (Matrix U) can be obtained by using the eigenvectors of L (Matrix V) as given by:

$$U_i = AV_i \quad (5)$$

The eifenfasecs are:

$$\text{eigenface} = [U_1, U_2, U_3,……. U_M] \quad (6)$$

Instead of using M eigenfaces, the highest m' $<=$ M is chosen as the eigenspace. Then the weight of each eigenvector $\omega_i$ to represent the image in the eigenface space, as given by:

$$\omega i = U_i^T (\Gamma - \Psi) , i=1,2,……, m' \quad (7)$$

$$\text{Weight matrix } \Omega = [\omega 1, \omega 2 …. \omega_{m'}]^T \quad (8)$$

$$\text{Average class projection } \Omega_\Psi = \frac{1}{X_i}\sum_{i=1}^{X_i} \Omega_i \quad (9)$$

The euclidean distance $\delta_i$ (8) is used to find out the distance between two face keys vectors and is given by:

$$\delta_i = \|\Omega - \Omega_{\Psi i}\| = \sum_{k=1}^{M}(\Omega_k - \Omega_{\Psi iK}) \quad (10)$$

The smallest distance is considered to be the face match score result.

## 3. RECENT WORK

Many existing face recognition researches use PCA (eigenfaces) for face recognition, we demonstrate some of the most recent work.

In [4], the authors introduce new measures, or scores, for symmetry of the face to assist in this analysis. The scores are computed using only the pixel values (and the weighted average of pixel values) of the images in the database. A 3D face database is used in their purposes to remove

25

unwanted errors in the symmetry calculation from issues present in 2D images, such as illumination. Using these scores, they perform statistical tests to compare the symmetry of the face in several subgroups of the database and then compare the face recognition results on the same subgroups. They show a significant difference in face symmetry scores between the men and women subgroups and compare the face recognition results. Then they divide the database into the most symmetric subjects and least symmetric subjects according to the symmetry scores and compare their face recognition results. They found that using symmetry in face recognition, by utilizing the average-half-face, is universally beneficial in their experiments. They have found a statistical significance between the face symmetry of men and women subjects in the 3D database as well as differences in face recognition accuracy. The least symmetric subjects produce higher face recognition accuracy than the most symmetric subjects when using the full face. However, face recognition accuracy is universally improved when utilizing the average-half-face in their experiments over the full face.

In [5], automated face recognition system has been developed in order to study the potential application for office door access control. The technique of eigenfaces based on the principle component analysis (PCA) and artificial neural networks have been applied into the system. The training images can be obtained either offline using in advance captured and cropped face images or online using the system's face detection and recognition training modules on the real frontal face images. The system has been able to recognize face at reasonable rate at the distance between 40 cm to 60 cm from the camera with the subject's head rotational angle is between -20 to +20 degrees. The experimental results have also confirmed the influences of illumination and pose toward face recognition system.

In [6], the authors employ principal component analysis (PCA) to extract features of face images, and sparse representation-based classification (SRC) algorithm is used to fulfil face recognition. experimental results show that whenever the optimal representation is sufficiently sparse it can be efficiently solved by convex optimization, which can be formulated as an l1-minimization problem. Moreover, the l1-minimization problem can be well solved by homotopy algorithm, thus the sparse coefficients are used to determine the corresponding object classes.

In [7], the authors propose a scheme which based on an information theory approach that decomposes face images into a small set of characteristic feature images called 'eigenfaces', which are actually the principal components of the initial training set of face images. Recognition is performed by projecting a new image into the subspace spanned by the eigenfaces ('face space') and then classifying the face by comparing its position in the facespace with the positions of the known individuals. The eigenface approach gives them efficient way to find this lower dimensional space. Eigenfaces are the eigenvectors which are representative of each of the dimensions of this face space and they can be considered as various face features. Any face can be expressed as linear combinations of the singular vectors of the set of faces, and these singular vectors are eigenvectors of the covariance matrices. The eigenfaces has proven the capability to provide the significant features and reduces the input size for neural network.

## 4. RESEARCH METHODOLOGY

Our work aims to improve the performance of PCA by decreasing the time of computation keeping same performance. We conduct three experiments using MatLab [8] each ensures one dimensional value for face recognition. The first experiment is used to adjust the best number of images for each individual to be used in the training set that gives a highest percentage of recognition. It is expected that the highest matching ratio is made by multiples of images in the training set for each person. This high number of images causes high computation time. In the second experiment we depend on the result of the first one and test 28 persons in the test database with six images for each person in the training database as given by experiment one. We change the threshold trying to make a decision of the best matching. In this experiment we observe the false acceptance rate FAR and the false rejection rate FRR for the algorithm and monitoring the time. In the third experiment we decrease the number of eigenvectors and consequently this decrease the time of computation. The results of this experiment give the same recognition results as the second experiment but with less time.

## 4.1. Database Description

We run our test database and training database depending on the face94 face database as follows: The face94 [2] is a face database constructed by Dr Libor Spacek. It is a part of a collection of facial images. This collection contains 4 folders of images face94, face95, face96 and grimace. We use the face94 which contains number 153 images with resolution of 180 by 200 pixels (portrait format), directories are female (20), male (113), male staff (20) contains images of male and female subjects in separate directories. The images are stored in 24 bit RGB, JPEG format and the used camera is S-VHS camcorder. The images are mainly of first year undergraduate students, so the majority of individuals are between 18-20 years old but some older individuals are also present. Some of them with glasses and the lighting is artificial, mixture of tungsten and fluorescent overhead.

Based on face94, we construct our own test and training databases. In test database we choose two images for 28 persons randomly and in the training database we choose six images for 40 persons. The images that we choose for the test database is differ from the training database. We choose six images in the training because the result of the first experiment shows that this number of images gives 100% recognition as we will show next section.

## 5. EXPERIMENTS AND RESULTS

We conduct three experiments each answer a question that help complete our research.

### 5.1 The First Experiment: Number of Images in the Training Database

In this experiment we aim to know what is number of images for a person that gives the best recognition percentage. Therefore, we test 19 persons in the test set with one image for each in the training set. Then, the number of images in training set is increased by a step of one. It shows that using 6 images in the training set for each person fulfils 100% recognition as shown in figure 1.

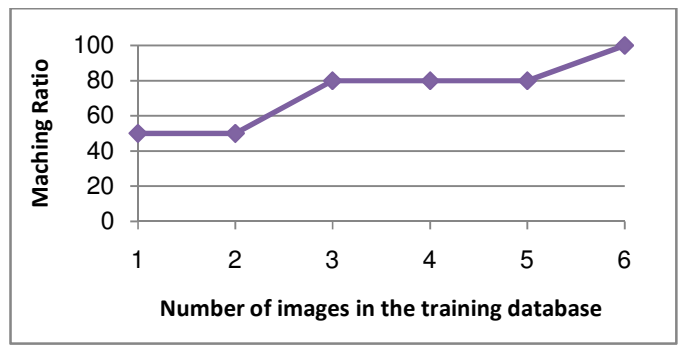

Figure 1 Matching ratio versus number of images in the training database

## 5.2. The Second Experiment: Which Threshold gives the Acceptable FAR, FRR?

In this experiment we test images for 28 persons in the test database against 40 persons each with 6 images in the training database. The threshold depends on the minimum euclidean distance. Euclidean distances between the projected test image and the projection of all cantered training images are calculated. Test image is supposed to have minimum distance with its corresponding image in the training database.

The curve in figure 2 (to simplify threshold value, we divide it by 1.0E+16) is used to determine the balance threshold value. The results show that equilibrium appears at a threshold equal to 4.50+E16 in which the FAR is equal to the FRR. But for more accuracy we choose 2.50+E16 as a threshold where FAR is nearly zero. In this experiment we include 230 eigenvectors and measure the time of each recognition process and it took between 1.9s to 2s. Figure 2 shows the trade off between FAR and FRR in the second experiment and figure 3 shows the DET curve [9] for this experiment.

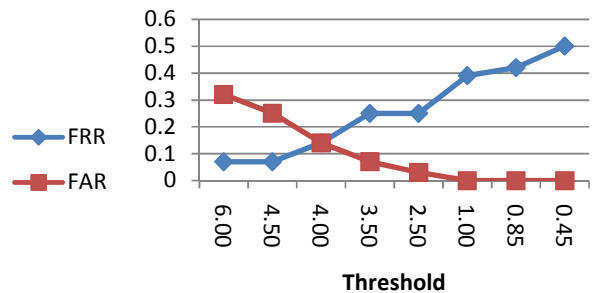

Figure 2: The tradeoff between FAR and FRR

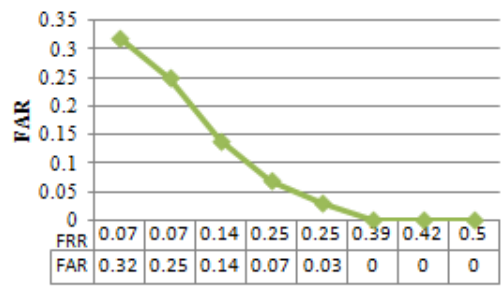

Figure 3 The DET Curve for the second experiment

### 5.3. The Third Experiment: Is Decreasing the Participated Eigenvectors Improve the Computational Performance of the Algorithm?

In this experiment we reduce the eigrnfaces for the PCA algorithm where eigenvalues are sorted and those who are less than a specified threshold are eliminated.

From the second experiment, we found that the eigenvectors ranging from -1.8949E-007 to 5.9288E+009. So in this experiment and after many trials, we adjust this threshold to 4.2201E+005 to reduce the number of the participated eigenvalues in the computation.
The second experiment is repeated with same parameters and instead of taking 230 eigenvectors we just use only 188 eigenvectors. The results of this experiment show a reduction of the computation time. Figure 4 and 5 show the results of the third experiment.

Applying our enhanced PCA algorithm in the third experiment, the time spent is in the range from 1.4s to 1.5s while keeping the same threshold value of 4.2201E+005at zero FAR.

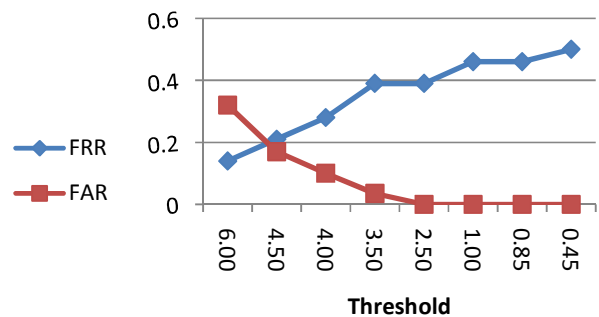

Figure 4: The tradeoff between FAR and FRR

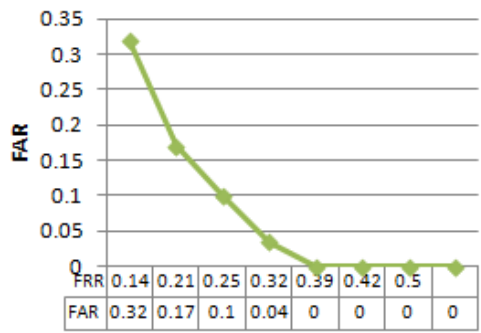

Figure 5 The DET Curve for the third experiment

## 6. DISCUSSION AND RECOMMENDATIONS

Increasing the number of images for each person in the training set to get best recognition rate causes long computational time which increased exponentially with the database size.

By comparing the results of the last two experiments, the enhanced algorithm gives a reduction of the computation time by 35% over the original PCA algorithm while keeps same performance.
We recommend the enhanced algorithm especially with the large database in where we need more computations.

## 7. CONCLUSIONS AND FUTURE WORK

In this work we aim to optimize the face recognition using eigenfaces algorithm by increasing the number of images in the training set and then reduce the computation time that may occur of this enhanced algorithm. Our enhanced algorithm reduces the participated eigenvectors in the algorithm to reduce the computation time. The enhanced algorithm gives the same performance results in less time of recognition as 35% of the recognition time of the original algorithm. Our experiments are conducted using the face94 face database and the algorithm is coded using MatLab.

Our research methodology is based on exercising the proposed modified PCA algorithm to decide the minimum number of images in the training set for individual that fulfill 100% of recognition, i.e., zero FAR. This experiment aims to neutralize the factor of the number of images in the training set while as this high training set number of images is best exercise our proposed algorithm for computational time.

It is also important to decide the best threshold value that fulfil the highest recognition rate, while as we use this value in the original PCA and enhanced one to compare the results.

For future work we want to repeat our experiment on larger databases. We also intent to conduct the same experiment using a different face database and compare the results with our current experiment to ensure the validity of our enhanced algorithm over different types and sizes of database. Other techniques to enhance the accuracy of the face recognition and decrease the false acceptance rate need to be compared.

International Journal of Artificial Intelligence & Applications (IJAIA), Vol.3, No.2, March 2012## REFERENCES

[1] A. S. Tolba; A.H. El-Baz; A.A. El-Harby:"Face Recognition: A Literature Review," World Academy of Science, Engineering and Technology 2006,
[2] M. Turk, A. Pentland, Eigenfaces for Recognition, Journal of Cognitive Neurosicence, Vol. 3, No. 1, 1991, pp. 71-86
[3] http://cswww.essex.ac.uk/mv/allfaces/
[4] Harguess, J.; Aggarwal, J.K.: "Is there a connection between face symmetry and face recognition?," *Computer Vision and Pattern Recognition Workshops (CVPRW), 2011 IEEE Computer Society Conference on* , vol., no., pp.66-73, 20-25 June 2011
[5] Ibrahim, R.; Zin, Z.M., "Study of automated face recognition system for office door access control application," *Communication Software and Networks (ICCSN), 2011 IEEE 3rd International Conference*, vol., no., pp.132-136, 27-29 May 2011
[6] Junying Gan; Peng Wang; , "A novel model for face recognition," *System Science and Engineering (ICSSE), 2011 International Conference on* , vol., no., pp.482-486, 8-10 June 2011
[7] Kshirsagar, V.P.; Baviskar, M.R.; Gaikwad, M.E., "Face recognition using Eigenfaces," *Computer Research and Development (ICCRD), 2011 3rd International Conference*, vol.2, no., pp.302-306, 11-13 March 2011
[8] Matlab, Version 7.0.0.19920 (R14), the MathWorks. Inc, Access date, June 22, 2010, from: http://www.mathworks.com/products/matlab/
[9] Martin A.; Doddington G.; Kamm T.; Ordowski M.; Przybocki M.: "The DET Curve in Assessment of Detection Task Performance," http://www.itl.nist.gov/iad/mig/publications/storage_paper/det.pdf
**Authors**

1- Dr Manal Abdullah received her PhD at computers and systems engineering, Faculty of engineering, Ain Shams University, Cairo, Egypt, 2002. She has experience in industrial computer networks and her research interest includes computer networks, performance evaluation, distributed DB, and bioinformatics. Currently she is assistant professor, Faculty of Computing and Information Technology FCIT, King Abdulaziz University, KAU, KSA.
2- Sahar Bo-Saeed grew up in Riyadh. She attended the University of KSU graduating with a major in Computer and Information Science (Honors in Computer Science). After graduation she worked as a teacher in public high school. She then went on to the Imam Moh. Bin Saud University as a teacher assistant, and she is completing her master degree in KSU.
3- Majda Wazzan born in Makkah, KSA. She is master student in computer Science at KAU. Her dissertation research in the area of Cloud Computing. She has bachelor's degree from KAU. She has worked for several years for KSU as a programmer and as a teacher assistant. Her research interests include design and analysis of computer algorithms, and web-based technologies. Now she worked for Deanship of IT at KAU.
31